\UseRawInputEncoding
\documentclass{article}
\pdfoutput=1

\usepackage[preprint]{w2wbev}

\usepackage[utf8]{inputenc} 
\usepackage[T1]{fontenc}    
\usepackage{hyperref}       
\usepackage{url}            
\usepackage{booktabs}       
\usepackage{amsfonts}       
\usepackage{nicefrac}       
\usepackage{microtype}      
\usepackage{xcolor}         

\usepackage{multirow}       
\usepackage{graphicx}
\usepackage{gensymb}
\usepackage{physics}
\usepackage{diagbox}
\usepackage{subfig}

\title{Window-to-Window BEV Representation Learning for Limited FoV Cross-View Geo-localization}

\author{%
		Lei~Cheng\\
	Department of Automation\\
	Southeast University\\
	Nanjing, Jiangsu 210019, China \\
	\texttt{leicheng@seu.edu.cn} \\
	\And
	Teng~Wang\\
	Department of Automation\\
	Southeast University\\
	Nanjing, Jiangsu 210019, China \\
	\texttt{wangteng@seu.edu.cn} \\
	\And
	Lingquan~Meng\\
	Department of Automation\\
	Southeast University\\
	Nanjing, Jiangsu 210019, China \\
	\texttt{menglingquan@seu.edu.cn} \\
	\And
	Changyin~Sun\\
	Department of Automation\\
	Southeast University\\
	Nanjing, Jiangsu 210019, China \\
	\texttt{cysun@seu.edu.cn}  \\
}

\begin{document}

\maketitle

\begin{abstract}
	Cross-view geo-localization confronts significant challenges due to large perspective changes, especially when the ground-view query image has a limited field of view with unknown orientation. To bridge the cross-view domain gap, we for the first time explore to learn a BEV representation directly from the ground query image. However, the unknown orientation between ground and aerial images combined with the absence of camera parameters led to ambiguity between BEV queries and ground references. To tackle this challenge, we propose a novel \textbf{W}indow-\textbf{t}o-\textbf{W}indow \textbf{BEV} representation learning method, termed \textbf{W2W-BEV}, which adaptively matches BEV queries to ground reference at window-scale. Specifically, predefined BEV embeddings and extracted ground features are segmented into a fixed number of windows, and then most similar ground window is chosen for each BEV feature based on the context-aware window matching strategy. Subsequently, the cross-attention is performed between the matched BEV and ground windows to learn the robust BEV representation. Additionally, we use ground features along with predicted depth information to initialize the BEV embeddings, helping learn more powerful BEV representations. Extensive experimental results on benchmark datasets demonstrate significant superiority of our W2W-BEV over previous state-of-the-art methods under challenging conditions of unknown orientation and limited FoV. Particularly, on the CVUSA dataset with FoV of 90 degree and unknown orientation, the W2W-BEV achieve an significant improvement from 47.24\% to 64.73 \%(+17.49\%) in R@1 accuracy.
\end{abstract}

\section{Introduction}
Cross-view geo-localization refers to the process of matching ground-level images against a database of satellite images annotated with accurate GPS labels to determine the geographic location of the query image. Due to the extensive coverage and cost-effectiveness of satellite imagery, this technology holds great prospect in the fields of robotic navigation and autonomous driving.
Existing researches~\cite{workman2015wide,lin2015learning,liu2019lending,toker2021coming,wang2021each} regard this task as an image retrieval problem, where the ground-level query image and the reference satellite image are mapped into the same embedding space for measuring their similarity.  However, the substantial viewpoint variation leads to significant disparities in appearance and spatial perception between the ground-level and satellite images, making the above representation learning extremely challenging.

To bridge the cross-view domain gap, previous works~\cite{shi2019spatial,shi2020optimal,zhang2023cross,shen2023mccg,zhu2023simple} primarily focus on designing a dual-branch convolutional neural network to extract features from different perspectives and aligning cross-view feature representations implicitly through deep metric learning. On the other hand, some methods~\cite{shi2019spatial,regmi2019bridging,fervers2023c} attempts to explicitly align different perspectives by transforming image from one domain into another domain in pixel~\cite{shi2020looking,shi2022accurate} or feature~\cite{shi2022beyond,shi2023boosting} space. For instance, polar transform~\cite{shi2019spatial} and projection transform~\cite{shi2022accurate} are developed to map aerial images to ground-level panoramas, whereas cross-view feature transformation is usually achieved by predicting the rotation and translation matrix ~\cite{shi2022beyond,shi2023boosting}. Despite significant improvements in recall rate, these geometry-based transformation can cause image distortion, leading to the loss of some valuable information. Furthermore, all of the above works make an assumption that the ground-level query image is a panorama and is aligned with the reference aerial image. Nevertheless, this premise may diverge from reality. In practical scenarios, ground images typically exhibit a limited Field of View~(FoV), with their orientations remaining unknown, which makes it more challenging to reduce disparities between cross-view images.  

\begin{figure}[t]
	\centering
	\includegraphics[width=1.0\textwidth]{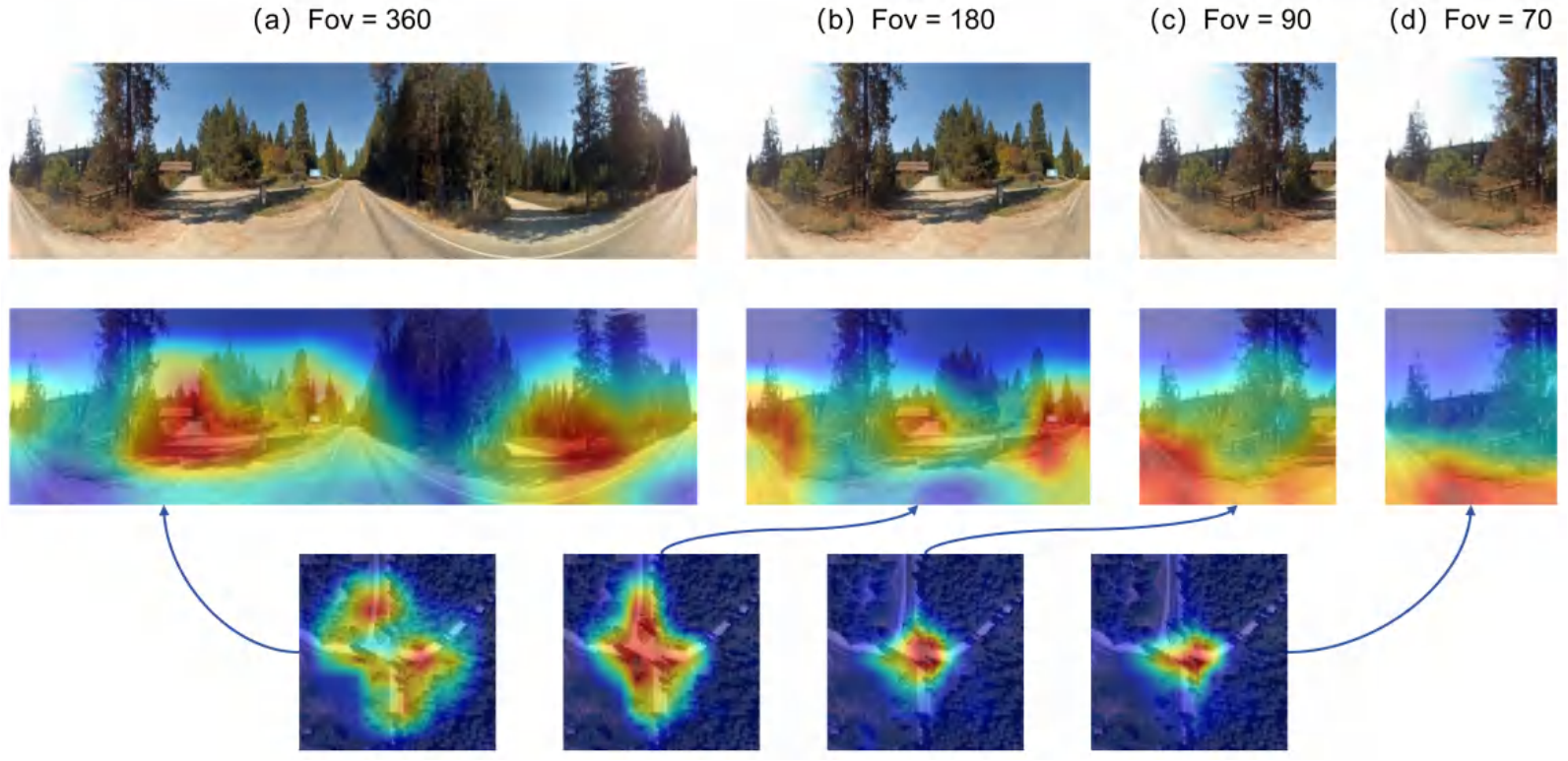}
	\caption{Illustration of ground images with unknown direction and different limited FoVs. The second and third rows respectively indicate the regions of interest corresponding to the BEV representations in ground-level images and aerial images. We can observe that BEV representations successfully extracts key spatial structure information from ground-level images. Furthermore, as the FoV of ground image increases, BEV representations can learn more extensive and effective information from the ground, corresponding to a broader range of interest in aerial images.}
	\label{fig:feature_vis}
	\vspace {-0.4cm}
\end{figure}

Recent researches~\cite{shi2020looking,yang2021cross,zhu2022transgeo,wang2023dehi,zhu2023simple,shi2022accurate,zhai2017predicting} have shifted their focus towards these more practical scenarios. Shi et al.~\cite{shi2020looking} propose a Dynamic Similarity Matching (DSM) model to estimate the relative orientation between cross-view features by computing their correlation along the azimuth angle axis. L2LTR~\cite{yang2021cross} and TransGeo~\cite{zhu2022transgeo} take advantage of the strong power of Vision Transformer~\cite{dosovitskiy2020image} in global modeling to learn robust global representations. DeHi~\cite{wang2023dehi} develop a decoupled hierarchical architecture, which is combined with the proposed part-level simplified bipartite matching (SimBiM) loss to achieve the orientation alignment of cross-view features. Although methods have shown significant improvements, they still struggle to meet the demands of practical applications.

To tackle this challenge, we propose a novel window-to-window Bird Eye's View (BEV) representation learning framework, dubbed W2W-BEV, to enhance real-world geo-localization with unknown orientation and limited FoV. We draw inspiration from methods~\cite{li2022bevformer, philion2020lift, li2023bevdepth} to projects ground-level image features into BEV space, aiming to bridge cross-view disparity in feature space while avoiding the loss of valuable information caused by image distortion. When the ground-level image has a limited FoV, we expect the BEV representations to capture  spatial structure information similar to that of reference aerial images by borrowing information from relevant ground-level image features, as shown in Figure~\ref{fig:feature_vis}. However, in practical scenarios where image orientation is unknown and camera parameters are missing, how to determine the positional correspondences between the ground-level image and the BEV embedding remains challenging. To tackle this issue, we develop a context-aware window matching strategy to determine the local correspondence between BEV embeddings and ground-level features in an adaptive manner. Additionally, we also introduce an initialization method for BEV embeddings using ground features and predicted depth information, which benefits the learning of BEV representation. Our extensive experimental results demonstrate that our method outperforms the state-of-the-art by a large margin under limited FoV and unknown orientation.

To summarize, our main contributions are as follows:
\begin{itemize}
	\item[$\bullet$] We propose a novel window-to-window BEV representation learning framework to enhance cross-view geo-localization under unknown orientation and limited FoV. This method learns BEV representation from ground-level image features to capture spatial structure information similar to that of aerial images, thereby reducing disparities between different perspectives.
	\item[$\bullet$] We develop a novel context-aware window matching strategy to determine the local correspondence between BEV embeddings and ground image features. When combined with the proposed BEV embedding initialization approach, it could effectively facilitate the learning of BEV representation.
	\item[$\bullet$] We conduct extensive experiments demonstrating that our method achieve significant improvements over the previous state-of-the-art methods. Particularly in scenarios with unknown orientation and limited viewpoints, our method outperform the state-of-the-art by a considerable margin.
\end{itemize}

\section{Related Work}
\paragraph{Feature extractor.}
Cross-view geo-localization can be regarded as an image retrieval task~\cite{workman2015wide,lin2015learning,vo2016localizing,hu2018cvm}. Due to the significant appearance and spatial differences caused by changes in viewpoint, it is crucial to extract robust global features. The widely used Spatial-aware Feature Aggregation module (SAFA) is proposed by Shi et al.~\cite{shi2019spatial}, which leverages attention to integrate spatial features. The method~\cite{shen2023mccg} provides a novel multiple classifier structure to acquire multiple feature representations for cross-view tasks. The geometric layout extractor and counterfactual-based Learning schema are introduced by work~\cite{zhang2023cross}, enabling the learning of robust spatial information. Recently, the vision transformer and its variants are widely utilized in recent works~\cite{yang2021cross,wang2023dehi,zhang2023cross,zhu2023simple} due to their robust global encoding capabilities. However, most of these methods rely heavily on clever module designs, and their performance significantly degrade when ground image perspectives are restricted, leading to a reduction in valuable information.

\paragraph{Viewpoint transformation.}
Because of the substantial challenges in implicitly aligning features from different viewpoints in the feature space through metric learning supervision, viewpoint transformation are garnering increasing attention from researchers. Methods~\cite{shi2019spatial, shi2020looking} employ the polar transformation to convert aerial images into ground-level images. However, this method assumes alignment between ground-level and aerial-level images orientations. Their further work~\cite{shi2022accurate} proposes the projection transformation to obtain a more realistic geometric relationship between ground-level panorama and satellite images. The Generative Adversarial Network(GAN) methods~\cite{regmi2019bridging,lu2020geometry} are utilized for generating images from different viewpoints, but they can easily introduce false image information that interfere with feature extraction. In contrast to the aforementioned pixel-level transformation, several methods~\cite{shi2022beyond,shi2023boosting} perform viewpoint transformation at the feature level. The approaches~\cite{shi2022beyond,shi2023boosting} project aerial images to ground-level in the feature space by predicting the unknown rotation matrix and translation vector.

\paragraph{Bird's-eye view.}
The bird's-eye view (BEV) is a commonly used representation of surrounding scenes because it clearly shows the position and scale of objects, making it suitable for various localization task. The LSS~\cite{philion2020lift} introduce a method for 2D-to-3D, which 'lift' 2D features to 3D voxel space based depth prediction~\cite{wang2019pseudo,you2019pseudo}. Constructing BEV features in a learnable manner, without requiring depth information, is proposed by BEVFormer~\cite{li2022bevformer}. Learning BEV representations is beneficial for reducing the viewpoint differences between ground-level and aerial-level images. Recently, some studies~\cite{fervers2023c,fervers2023uncertainty,wang2024fine} have started applying the BEV methods to cross-view geo-location tasks as well. In the case where camera parameters are available, The work~\cite{fervers2023uncertainty} adopts a method similar to BEVFormer~\cite{li2022bevformer} to learn BEV representations from ground features. When camera parameters are not available, The C-BEV\cite{fervers2023c} employs attention operations to project image features onto a polar coordinate system, followed by mapping them from the polar coordinate system to cartesian representation using bilinear resampling to obtain BEV representation. The spherical transformation~\cite{wang2024fine} is utilized for ground panorama images, projecting them onto the BEV perspective through geometric transformations, which still distort the images and lead to information loss. However, common learnable BEV methods typically require finding the reference of the BEV query in ground image features, which often involves using camera parameters to determine their physical-world positions. Therefore, when facing missing camera parameters, unknown orientation and limited FoV of ground-level images, we propose a window-to-window method based on a context-aware window matching strategy to effectively learn the BEV representation.
\begin{figure*}[t]
	\centering
	\includegraphics[width=1.01\textwidth]{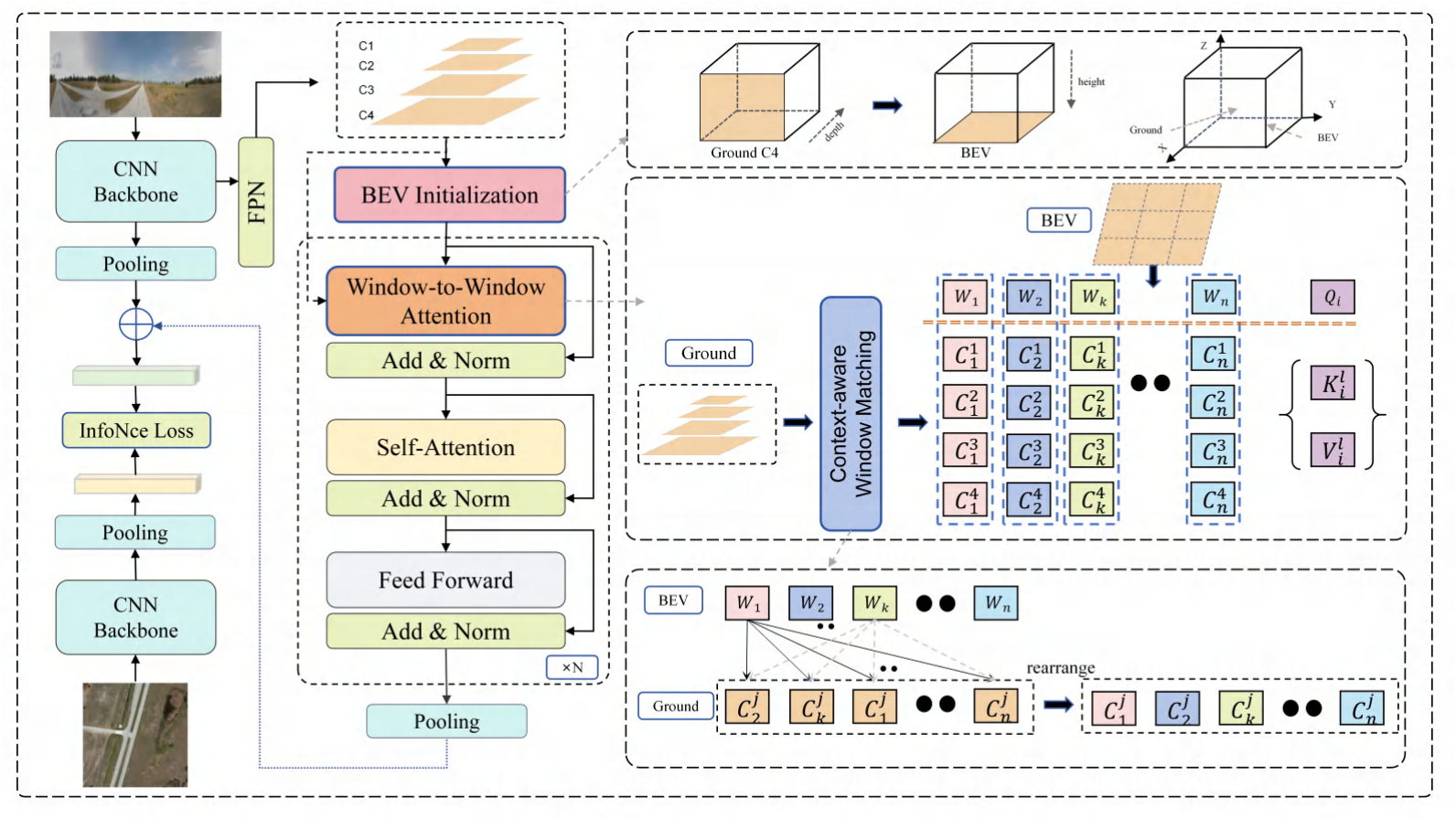}
	\caption{The overview of proposed method. To facilitate the learning of BEV representations, we utilize ground feature C4 from multi-scale features to predict it depth probability. This extends the 2D features into 3D, which are then compressed along $H$ dimension to generate initial BEV embeddings.}
	\label{fig:model_structure}
\end{figure*}

\section{Methodology}
We propose a window-to-window BEV representation learning method to align cross-view images in feature space for enhancing cross-view geo-localization under limited FoV and unknown orientation. Figure~\ref{fig:model_structure} illustrates the overall framework of our model. The following sections will specifically introduce each component of our method. 

\subsection{Multi-Scale Feature Extraction}
We harness ConvNeXt\_base~\cite{liu2022convnet} as our backbone to extract features from input ground-level images, which has demonstrated powerful feature extraction capabilities in recent works~\cite{fervers2023uncertainty,deuser2023sample4geo}. In convolutional neural networks, features from deeper layers capture rich semantic information, but they are in low resolution, limiting the geometric information they contain. Therefore, while these deep features are beneficial for BEV representation learning, they are not optimal due to the importance of spatial details captured by earlier layers. Therefore, we exploit the multi-scale features extracted from the ground-level images for BEV representation learning. Similar to the methods~\cite{lin2017feature,guo2020augfpn}, we extract the intermediate features of four stages in the ConvNeXt network, as shown in Figure~\ref{fig:model_structure}. Then, we sequentially apply upsampling and addition operations to obtain the final multi-scale features, denoted as $C_1$, $C_2$, $C_3$, and $C_4$.

\subsection{Window-to-Window BEV representation Learning}
Existing BEV-related works including LSS~\cite{philion2020lift} and BEVFormer~\cite{li2022bevformer} typically leverage camera parameters to establish point-to-point correspondence between BEV space and image features. However, this is not applicable for our task, since mainstream datasets like CVUSA~\cite{zhai2017predicting} and CVACT~\cite{liu2019lending} lack annotated camera parameters. Furthermore, when the ground-level query image suffers from unknown orientation and limited FoV, the spatial correspondence between BEV embedding and ground images becomes more challenging to determine. To enable a learning-based approach for obtaining BEV representations under these extreme conditions, we develop a context-aware matching strategy to achieve window-to-window correspondence as a substitute for point-to-point correspondence. Specifically, our window-to-window BEV learning procedure is mainly divided into three steps. Firstly, we extract depth information from the multi-scale features to build an initial BEV embedding. Then, we adopt a context-aware window matching strategy to align windows which can help establish a rough correspondence. Finally, we perform cross-attention based on the corresponding windows to learn BEV representations from ground-level images.

\paragraph{BEV embedding Initialization.}
The street view images are captured by ground cameras, which compress depth information during the formation of 2D images. On the other hand, for aerial images or BEV representation, height information in physical space is also compressed. However, both ground images and BEV representation should retain features about the horizontal dimensions in physical space. Therefore, we believe that using ground images to generate a reliable initial state for BEV embedding is beneficial for subsequent attention-based BEV representation learning. This is achieved by projecting ground features back to 3D space based on the predicted depth information. Figure~\ref{fig:model_structure} illustrates the procedure of initializing our BEV embedding. Specifically, we select the $C_4$ feature from the multi-scale features of ground images because it integrates features from different stages, providing rich semantic information while preserving geometric information. We first use a fully connected layer and softmax to predict the depth probability distribution of ground features, projecting them from 2D to 3D. Then, we compress the height information of 3D features using max pooling to obtain the initial value of BEV embedding.

\paragraph{Context-aware window matching strategy.}
To establish the correspondence between windows, the BEV embeddings and ground-level features are segmented into several windows, respectively. Specifically, for aerial images, we divide them into N windows in a grid-like manner, defined as $W_i~(i = 1,2,...,N) \in \mathbb{R}^{\frac{H_a}{\sqrt{N}} \times \frac{W_a}{\sqrt{N}} \times C}$. For ground-level images, we divide the image features into N windows along the horizontal direction, defined as $C_j^l~(l = 1,2,3,4) (j = 1,2,...,N) \in \mathbb{R}^{H_g \times \frac{W_a}{N} \times C}$, where $l$ denotes the multi-scale features level, $(H_a, W_a)$ and $(H_g, W_g)$ denote the height and width of the aerial image and ground image, and $C$ represents the channel number of features. Then, we search for the best matching ground windows for each BEV embedding window based on context-aware window matching strategy. To be specific, We perform global average pooling on each window to obtain a feature representation. Then, we calculate the similarity between the BEV embedding window and each ground-level feature window. We consider the ground-level window with the highest similarity as the corresponding window for the BEV window. Taking $C_1$ as an example, there are $N$ ground-level window features, as $C_j^1$. For any given aerial window feature $W_k$, we define a matching window pair as $(W_k, C_k^1)$, where $C_k^1$ satisfies:
\begin{equation}
	C_k^1=\arg \max _{C_j^1} \operatorname{Corr}\left(\operatorname{Avg}(W_k), \operatorname{Avg}(C_j^1)\right),
	\setlength\belowdisplayskip{1.5pt}
\end{equation}
where $\operatorname{Corr}$ represents the correlation calculation, which is a dot-product operation, and Avg denotes global average pooling operation. Therefore, unlike point-to-point relationships determined by geometric positions, the window-to-window approach establishes local correspondences through perceiving contextual features within windows.
\paragraph{BEV Encoder.} 
To learn a robust BEV representation, we introduce cross-attention~\cite{lin2022cat} and self-attention~\cite{vaswani2017attention} based on the established window correspondence. The vanilla multi-head attention operation is as follows:
\begin{equation}
	\operatorname{Attention~}(Q, K, V)=\operatorname{softmax}\left(\frac{Q K^{T}}{\sqrt{C}} V\right),
\end{equation}
\begin{equation}
	\text {head }_{\mathrm{i}}= \operatorname{Attention~}\left(Q W_{i}^{Q}, K W_{i}^{K}, V W_{i}^{V}\right), \\
\end{equation}
\begin{equation}
	\operatorname{MultiHead~}(Q, K, V)=\operatorname{Concat~}\left(\text { head }_{1}, \ldots, \text { head }_{\mathbf{h}}\right) W^{O}, \\
\end{equation}
where Q, K, and V respectively represent Query, Reference, and Key,  $\operatorname{Concat~}$ denotes concatenation along the channel dimension, and $W_{i}^{Q}$, $W_{i}^{K}$, $W_{i}^{V}$, and $W^{O}$ represent the weights of different fully connected layers. As shown in Figure~\ref{fig:model_structure}, within their respective corresponding windows, the BEV embedding generates Query through a linear layer, while the ground image features provide Key and Value, allowing them to interact and learn effective BEV representation. In addition to the window-to-window attention, since the BEV construction process is local, self-attention is employed for global interaction and a feedforward neural network is used to provide non-linearity. The whole encoding procedure could be described as follows:
\begin{equation}
	W_i = \sum_{l=1}^{4} \operatorname{MultiHead~}(W_i,C_i^l,C_i^l),
\end{equation}
\begin{equation}
	W_i = \operatorname{MultiHead~}(W_i,W, W),	
\end{equation}
\vspace {2pt}
\begin{equation}
	W_i = \operatorname{FFN~}(W_i),
\end{equation}
where W denotes the whole BEV windows, and $\operatorname{FFN}$ represents feedforward neural network.
\section{Experiment}
We first introduce two benchmark datasets, evaluation metrics, and implement details of our method. Then, we compare the performance of our method with state-of-the-art works across datasets. Finally, we present our ablation experiments and qualitative analysis to validate the effectiveness of our approaches.

\subsection{Datasets and Evaluation Metrics}
\paragraph{CVUSA and CVACT Datasets.} The benchmarks CVUSA\cite{zhai2017predicting} and CVACT\cite{liu2019lending} are commonly used in cross-view geo-localization. Both datasets consist of pairs that involve one-to-one matching, including a panoramic image and a directionally aligned aerial image. The CVUSA dataset provides 35,532 pairs of images for training and 8,884 pairs for testing. The CVACT dataset includes 35,532 pairs for training and 8,884 pairs for validation (denoted as CVACT\_val). Additionally, CVACT offers a larger version of the dataset for testing (denoted as CVACT\_test), containing 92,802 pairs of images. Our experiments simulate real-world scenarios where ground image orientations are unknown and viewpoints are limited by randomly shifting and cropping the panoramic images from both the CVUSA and CVACT datasets.

\paragraph{Evaluation Metrics.}
we employ recall accuracy at recall-k, noted as R@k, as our primary evaluation metric, and report recalls at top-1, top-5, top-10 and top 1\%. Specifically, for a given query image, we consider the retrieval successful in terms of the R@K metric when the positive sample is among the top K retrieved samples. 

\subsection{Implementation Details}
We crop ground-level panoramic images to 192 $\times$ 768 and aerial images to 384 $\times$ 384 on both datasets. During training, we randomly translate panoramic images and further crop them along the width direction to simulate real scenarios with unknown orientation and limited FoV. We use ConvNext\_base as the backbone model. Similar to the approach~\cite{deuser2023sample4geo}, we use the InfoNCE loss~\cite{oord2018representation,radford2021learning} for metric learning, aiming to make the distances between matched image pairs closer and farther for non-matched image pairs in feature space. We split the BEV embeddings and ground features into four windows. In our main experiment, the number of blocks in our BEV encoder is 3 and the number of attention heads is 4, and the size of the BEV embeddings is set to 28 $\times$ 28. Our experiments are based PyTorch, and our models are trained using the AdamW~\cite{loshchilov2017decoupled} optimizer with a cosine learning rate schedule. If not specified, all the experiments are conducted on 4 24GB NVIDIA GeForce RTX 4090 GPUs, with the initial learning rate of 0.0001 and batch size of 16. Additionally, we employ the same data mining approach as Sample4Geo~\cite{deuser2023sample4geo}. To enhance the superiority of our method, we reproduce Sample4Geo performance under unknown orientation and limited FoV by using the same batch size and learning rate as ours. Below, we demonstrate the performances of our method across multiple experimental settings.
\begin{table}[t]
	\caption{The performance comparison with state-of-the-art methods on the CVUSA and CVACT datasets with unknown orientation. $\dagger$ denotes which models are using the polar transformation.}
	\label{Unalign Cross-View Geo-Localization}
	\centering
	\scalebox{0.75}{
		\begin{tabular}{l|cccc|cccc|cccc}
			\toprule
			\multirow{2}*{Approach} & \multicolumn{4}{c}{CVUSA} & \multicolumn{4}{c}{CVACT\_val} & \multicolumn{4}{c}{CVACT\_test} \\
			\cline{2-13}
			& R@1 & R@5 & R@10 & R@1\% & R@1 & R@5 & R@10 & R@1\% & R@1 & R@5 & R@10 & R@1\% \\
			\midrule
			CVM-Net~\cite{hu2018cvm}                  & 16.25 & 38.86 & 49.41 & 88.11 & 13.09 & 33.85 & 45.69 & 81.80 & - & - & - & - \\
			CVFT~\cite{shi2020optimal}                & 23.38 & 44.42 & 55.20 & 86.64 & 26.79 & 46.89 & 55.09 & 81.03 & - & - & - & - \\
			$\dagger$DSM~\cite{shi2020looking}        & 78.11 & 89.46 & 92.90 & 98.50 & 72.91 & 85.70 & 88.88 & 95.28 & - & - & - & - \\
			$\dagger$DeHi~\cite{wang2023dehi}         & 82.38 & 93.53 & 96.30 & \textbf{99.36} & 77.94 & 90.62 & 93.26 & 97.65 & - & - & - & - \\
			Sample4geo~\cite{deuser2023sample4geo}  & 90.80 & 97.14 & 98.05 & 99.10 & 81.27 & 89.98 & 91.68 & 95.70 & 64.49 & 82.59 & 85.68 & 95.78 \\ 
			\textbf{Ours} & \textbf{93.22} & \textbf{97.99} & \textbf{98.72} & 99.35 & \textbf{83.09} & \textbf{91.20} & \textbf{92.98} & \textbf{96.45} & \textbf{65.91} & \textbf{84.32} & \textbf{87.28} & \textbf{96.55}                               \\
			\bottomrule
	\end{tabular}}
\end{table}

\subsection{Comparing with State-of-the-art Models} 
We compared our method with some state-of-the-art methods on CVUSA, CVACT\_val, and CVACT\_test datasets. Our method first introduce a window-to-window way to learn BEV representation based context-aware window matching strategy to solve cross-view geo-localization with orientation unknown and limited FoV. 

\paragraph{Cross-View Geo-Localization With Unknown Orientation.}
We first evaluated our method on the cross-view datasets CVUSA and CVACT with unknown Orientation. We compare our method with some state-of-the-art methods, including CVM-Net~\cite{hu2018cvm}, CVFT~\cite{shi2020optimal}, DSM~\cite{shi2020looking}, DeHi~\cite{wang2023dehi}, and Sample4Geo~\cite{deuser2023sample4geo}. Some method of above~\cite{shi2020looking,wang2023dehi} employ polar transformations to reduce viewpoint discrepancies and extract effective global features through the design of subtle modules. On the other hand, Sample4Geo~\cite{deuser2023sample4geo} takes a data-centric approach, proposing two effective data mining methods. In contrast, our proposed method reduces the differences between viewpoints by learning BEV representation from ground image features. In Table~\ref{Unalign Cross-View Geo-Localization}, when the orientation is unknown, compared to the previously state-of-the-art methods, our approach improve from 90.80\% to 93.22\%(+2.42\%) on CVUSA, from 81.27\% to 83.09\%(+1.82\%) on CVACT\_val, and from 64.49\% to 65.91\%(+1.42\%) on CVACT\_test. This indicates that even without viewpoint alignment, our approach can still extract effective local and spatial information.

\begin{table}[h]
	\caption{The performance comparison with state-of-the-art methods on the CVUSA dataset with unknown orientation and limited FoV. $\dagger$ denotes which models are using the polar transformation.}
	\label{Fov}
	\centering
	\scalebox{0.67}{
		\begin{tabular}{l|c|cccc|cccc|cccc}
			\toprule
			\multirow{2}*{Dataset} & \multirow{2}*{Approach} & \multicolumn{4}{c}{Fov = 180\degree} & \multicolumn{4}{c}{Fov = 90\degree} & \multicolumn{4}{c}{Fov = 70\degree} \\
			\cline{3-14}
			& & R@1 & R@5 & R@10 & R@1\% & R@1 & R@5 & R@10 & R@1\% & R@1 & R@5 & R@10 & R@1\% \\
			\midrule
			\multirow{9}*{CVUSA} & CVM-Net~\cite{hu2018cvm}                & 7.38 & 22.51 & 32.63 & 75.38 & 2.76 & 10.11 & 16.74 & 55.49 & 2.62 & 9.30 & 15.06 & 21.77 \\
			& CVFT~\cite{shi2020optimal}                                   & 8.10 & 24.25 & 34.47 & 75.15 & 4.80 & 14.84 & 23.18 & 61.23 & 3.79 & 12.44 & 19.33 & 55.56 \\
			& $\dagger$DSM~\cite{shi2020looking}                           & 48.53 & 68.47 & 75.63 & 93.02 & 16.19 & 31.44 & 39.85 & 71.13 & 8.78 & 19.90 & 27.30 & 61.20 \\	
			& GAL~\cite{rodrigues2022global}                               & 48.91 & 69.87 & 78.50 & 95.68 & 22.54 & 44.36 & 54.17 & 83.59 & 15.20 & 32.86 & 42.06 & 75.21 \\
			& $\dagger$L2LTR~\cite{yang2021cross}                          & 56.69 & 80.86 & 87.75 & 98.01 & 26.92 & 50.49 & 60.41 & 86.88 & 13.95 & 33.07 & 43.86 & 77.65 \\
			& TranGeo~\cite{zhu2022transgeo}                               & 58.22 & 81.33 & 87.66 & 98.13 & 30.12 & 54.18 & 63.96 & 89.18 & - & - & - & - \\	
			& $\dagger$DeHi~\cite{wang2023dehi}                            & 60.42 & 81.81 & 88.16 & 98.03 & 31.53 & 55.13 & 65.57 & 90.84 & - & - & - & - \\
			& $\dagger$ArcGeo~\cite{shugaev2024arcgeo}                                              & - & - & - & - & 44.18 & 70.33 & 78.84 & - & - & - & - & - \\
			& Sample4geo~\cite{deuser2023sample4geo}                       & 83.51 & 94.54 & 96.23 & 98.83
			& 47.24 & 71.24 & 79.14 & 93.96 & 36.77 & 62.20 & 71.92 & 92.06 \\
			& \textbf{Ours}                                                & \textbf{86.55} & \textbf{95.63} & \textbf{97.13} & \textbf{98.95} & \textbf{64.75} & \textbf{84.66} & \textbf{89.37} & \textbf{96.88} & \textbf{43.33} & \textbf{68.73} & \textbf{77.41} & \textbf{93.78} \\
			\midrule
			\multirow{5}*{CVACT\_val} & CVM-Net~\cite{hu2018cvm}                                   & 3.94 & 13.69 & 21.23 & 59.22 & 1.47 & 5.70 & 9.64 & 38.05 & 1.24 & 4.98 & 8.42 & 34.74 \\
			& CVFT~\cite{shi2020optimal}                                   & 7.13 & 18.47 & 26.83 & 63.87 & 1.85 & 6.28 & 10.54 & 39.25 & 1.49 & 5.13 & 8.19 & 34.59 \\
			& $\dagger$DSM~\cite{shi2020looking}                           & 49.12 & 67.83 & 74.18 & 89.93 & 18.11 & 33.34 & 40.94 & 68.65 & 8.29 & 20.72 & 27.13 & 57.08 \\	
			& GAL~\cite{rodrigues2022global}                               & 49.93 & 68.48 & 77.16 & 93.01 & 26.05 & 49.23 & 59.26 & 85.60 & 14.17 & 32.96 & 43.24 & 77.19 \\
			& Sample4geo~\cite{deuser2023sample4geo}                       & 58.36 & 77.25 & 83.01 & 93.73
			& 26.70 & 50.69 & 60.96 & 85.05 & 18.37 & 38.88 & 48.98 & 78.55 \\ 
			& \textbf{Ours} &\textbf{63.17} & \textbf{81.09} & \textbf{86.10} & \textbf{94.89} & \textbf{32.22} & \textbf{57.49} & \textbf{66.85} & \textbf{88.16} & \textbf{24.28} & \textbf{47.70} & \textbf{58.16} & \textbf{84.87} \\
			\midrule
			CVACT\_test  & Sample4geo~\cite{deuser2023sample4geo}          & 32.59 & 55.64 & 63.30 & 92.50
			& 9.82 & 23.99 & 31.72 & 82.70 & 5.09 & 14.79 & 20.81 & 75.38 \\ 
			& \textbf{Ours} & \textbf{37.88} & \textbf{62.35} & \textbf{69.67} & \textbf{94.99} & \textbf{12.67} & \textbf{30.14} & \textbf{38.85} & \textbf{88.56} & \textbf{7.58} & \textbf{21.09} & \textbf{28.89} & \textbf{85.16} \\
			\bottomrule
	\end{tabular}}
\end{table}	

\paragraph{Cross-View Geo-Localization With Unknown Orientation and Limited FoV.}
In a more realistic localization scenario with unknown orientation and limited FoV, we compare our method with state-of-the-art approaches on CVUSA and CVACT datasets. Table~\ref{Fov} demonstrates the performance of our method across fields of view 180\degree, 90\degree, and 70\degree, and compares it with previous approaches. Our method achieves significant improvements in situations where the FoV is limited. When the FoV is 180\degree on the CVUSA, our method achieve an accuracy of 86.55\% in the R@1, showing a certain improvement of 3.04\% over the previous state-of-the-art method, which has an accuracy of 83.51\%. Particularly noteworthy is the remarkable improvement of our model at the 90\degree FoV, rising from 47.24\% to 64.75\% (+17.51\%), which indicates that BEV representations could learn effective spatial structure features when the effective information of ground image is severely reduced. 
Even when the FoV is reduced to a particularly small number of 70 \degree, our method still surpasses state-of-the-art method by 6.56\%. On the CVACT\_val, improvements of 4.81\%, 5.52\%, and 5.91\% are achieved at FoV of 180\degree, 90\degree, and 70\degree, respectively. Notably, substantial progress is also made on the larger CVACT\_test dataset, with an increase from 32.59\% to 37.88\% (+5.29\%) at FoV of 180\degree.

\begin{figure}[t]
	\centering
	\includegraphics[width=1.02\textwidth]{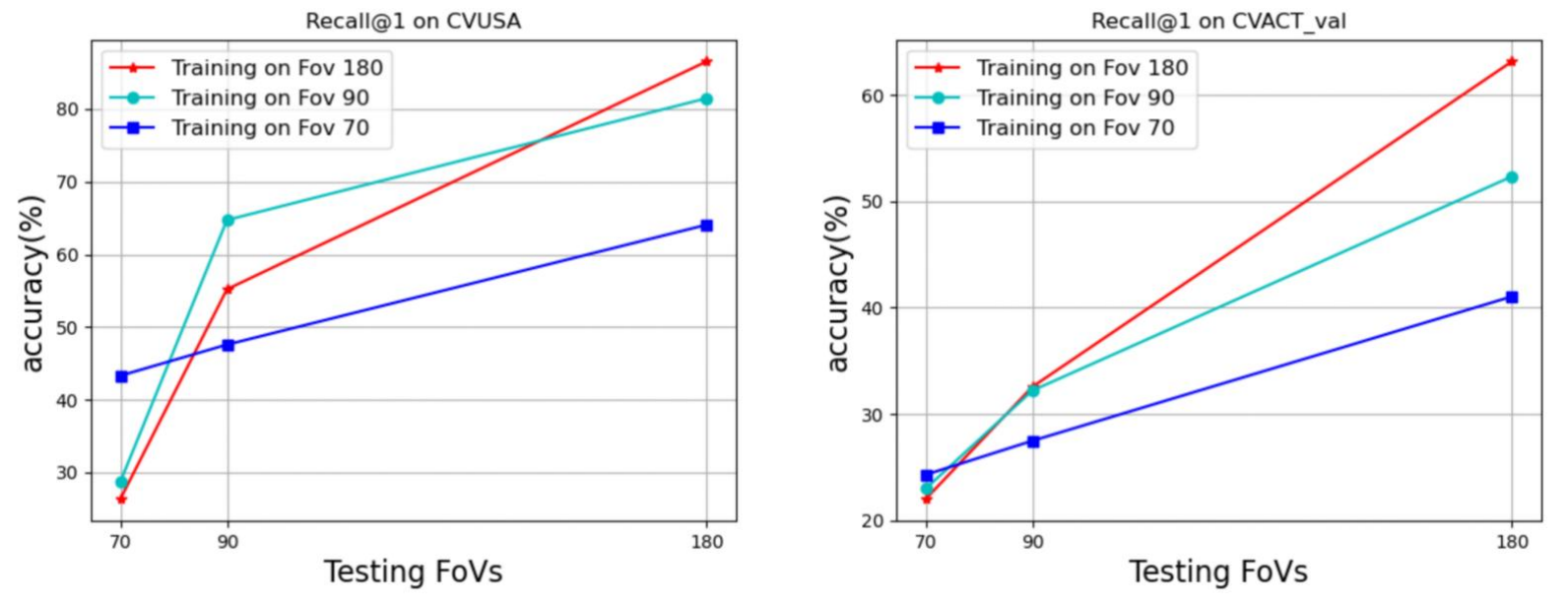}
	\caption{Recall performance at top-1 of our model with different training and testing of FoVs.}
	\label{fig:fovs}
\end{figure}

\paragraph{Training and Testing on Different FoVs.}
In real-world scenarios, the camera's FoV may not align with used during training. Therefore, evaluating the performance of a trained model across different viewpoints holds practical significance. Hence, we investigate the performance generalization of models tested across various FoVs. As shown in Table~\ref{angle generalization}, we present the performance of our model trained at 180, 90, and 70 degrees, respectively, when tested at different degrees. As depicted in Fig~\ref{fig:fovs}, we observe the improvement of models performance as the FoV of the test dataset increases, owing to the provision of more effective information. Furthermore, for testing at a particular FoV, model trained at the same FoV tends to perform the best, which underscores the optimal performance achieved when training and testing distributions align.
\begin{table}[h]
	\caption{Illustration our method for training and testing on different FoVs.}
	\label{angle generalization}
	\centering
	\scalebox{0.7}{
		\begin{tabular}{l|c|cccc|cccc|cccc}
			\toprule
			\multirow{2}*{Dataset} & \multirow{2}*{\diagbox{train}{test}} & \multicolumn{4}{c}{Fov = 180\degree} & \multicolumn{4}{c}{Fov = 90\degree} & \multicolumn{4}{c}{Fov = 70\degree} \\
			\cline{3-14}
			& & R@1 & R@5 & R@10 & R@1\% & R@1 & R@5 & R@10 & R@1\% & R@1 & R@5 & R@10 & R@1\% \\
			\midrule
			\multirow{3}*{CVUSA} & 180\degree               & 86.55 & 95.63 & 97.13 & 98.95 & 55.19 & 73.62 & 79.45 & 90.50 & 26.31 & 46.68 & 54.85 & 75.38 \\
			& 90\degree                                     & 81.54 & 94.05 & 96.17 & 99.04 & 64.75 & 84.66 & 89.37 & 96.88 & 28.77 & 51.99 & 61.04 & 82.33 \\
			& 70\degree                                     & 64.09 & 85.27 & 90.72 & 98.28 & 47.52 & 73.04 & 80.99 & 95.37 & 43.33 & 68.73 & 77.41 & 93.78 \\	
			
			\midrule
			\multirow{3}*{CVACT\_val} & 180\degree                                   & 63.17 & 81.09 & 86.10 & 94.89 & 32.56 & 54.99 & 64.15 & 85.51 & 22.04 & 42.83 & 52.06 & 77.90 \\
			& 90\degree                                   & 53.05 & 74.63 & 81.49 & 93.88 & 32.22 & 57.49 & 66.85 & 88.16 & 23.28 & 46.27 & 55.89 & 82.95 \\
			& 70\degree                                   & 41.06 & 65.14 & 73.50 & 91.84 & 27.46 & 51.64 & 62.06 & 86.71 & 24.28 & 47.70 & 58.16 & 84.87 \\	
			\bottomrule
	\end{tabular}}
\end{table}

\subsection{Ablation Study} 
We demonstrate the impact of our approach and several parameters in the proposed method. First, as shown in Table~\ref{table:depth}, we verify that initializing the BEV embeddings using ground features combined with predicted depth information significantly improves the model's performance.This is because ground-level and aerial images share visual information in the horizontal dimension, and using depth information as the initial value for the vertical dimension of BEV embeddings benefits the subsequent learning of BEV representations. 
\begin{table}[h]
	\caption{The effects of initializing BEV embedding with ground depth information are compared.}
	\label{table:depth}
	\centering
	\scalebox{0.72}{
		\begin{tabular}{l|cccc|cccc|cccc}
			\toprule
			\multirow{2}*{Approach} & \multicolumn{4}{c}{Fov = 180\degree} & \multicolumn{4}{c}{Fov = 90\degree} & \multicolumn{4}{c}{Fov = 70\degree} \\
			\cline{2-13}
			& R@1 & R@5 & R@10 & R@1\% & R@1 & R@5 & R@10 & R@1\% & R@1 & R@5 & R@10 & R@1\% \\
			\midrule
			Ours(w/o BEV initial) & 85.32 & 95.40 & 97.11 & 99.28 & 62.48 & 82.96 & 88.69 & 97.58 & 41.21 & 67.60 & 76.37 & 94.02\\
			\textbf{Ours}                                                & \textbf{86.55} & \textbf{95.63} & \textbf{97.13} & \textbf{98.95} & \textbf{64.75} & \textbf{84.66} & \textbf{89.37} & \textbf{96.88} & \textbf{43.33} & \textbf{68.73} & \textbf{77.41} & \textbf{93.78} \\
			\bottomrule
	\end{tabular}}
\end{table}

\begin{figure}[t]
	\centering
	\includegraphics[width=1.0\textwidth]{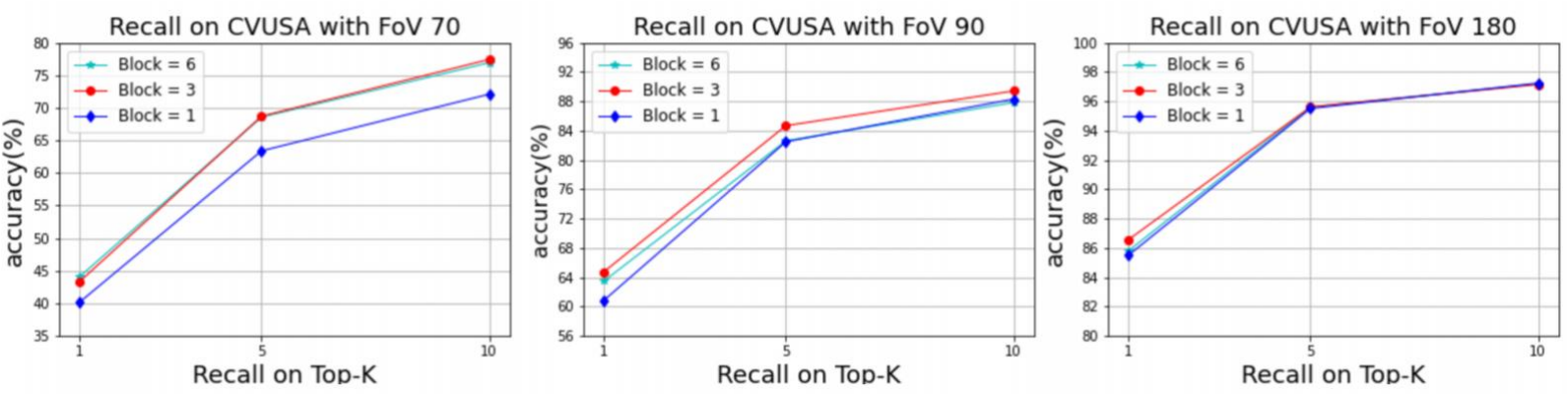}
	\caption{The effect of the number of BEV encoder blocks on Recall accuracy with different FoVs is compared.}
	\label{fig:block}
\end{figure}

\begin{figure}[ht]
	\centering
	\includegraphics[width=1.0\textwidth]{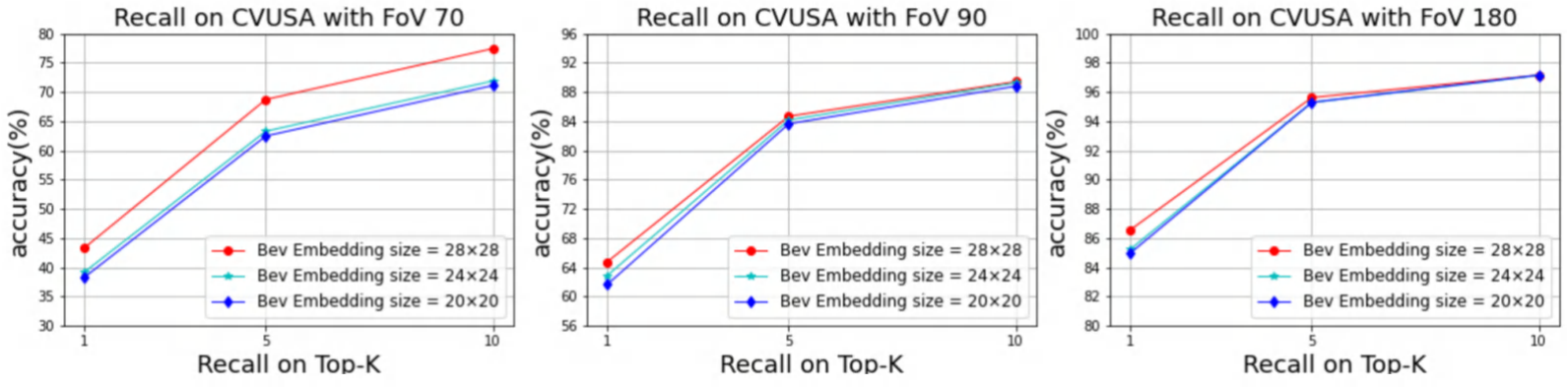}
	\caption{The effect of the size of BEV embedding on Recall accuracy with different FoVs is compared.}
	\label{fig:embedding}
\end{figure}

Next, we further compare the impact of the number of blocks in the BEV encoder and the size of the BEV embeddings. As shown in Figure~\ref{fig:block}, when the number of blocks is reduced to 1, the model's performance decreases across different FoVs. However, it is noteworthy that increasing the number of blocks to 6 does not improve the model's performance. We speculate that this is because, under limited FoV, the effective information is reduced, and simply increasing the number of blocks may lead to overfitting.
Figure~\ref{fig:embedding} shows the impact of the BEV embedding size. As the size of the BEV embeddings increases, the model's performance consistently improves. This improvement is due to the larger BEV embeddings facilitating the learning of more detailed information from the ground features. However, it is important to note that this also results in increased computational overhead.
\subsection{Analysis} 
To explain why our method can achieve excellent performance, we visualize the BEV representations of the regions of interest in both the ground view and the aerial view images by employing the Grad-cam~\cite{selvaraju2017grad} method. As shown in Figure~\ref{fig:feature_vis}, the BEV representations learned by our method align the regions of interest in the ground-level and aerial-level images, which demonstrates that the BEV representation can successfully learn spatial structural information similar to that of the aerial view from ground features. Additionally, we observe that as FoV of ground-level images increases, the corresponding regions in the aerial image for the BEV representation also increase, which indicates that the increase in ground-level information facilitates the learning of BEV representations.

\section{Conclusion and Limitation}
In the task of cross-view geo-localization with unknown orientation and limited field of view, we propose a novel Window-to-Window BEV representation learning method, termed W2W-BEV, which adaptively match BEV queries to ground reference at window-scale. By learning BEV representations through ground features, our method effectively reduces differences between perspectives. Our approach achieves significant improvements, especially when the FoV is less than 180 degree. However, One main limitation of our method is that the additional introduction of BEV embeddings increases the GPU memory and computational requirements. Therefore, in future work, we hope to propose a low-memory consumption model for cross-view geo-localization. Furthermore, we believe that more accurate depth information will help further optimize the model's performance. Therefore, in future work, designing a precise depth prediction module or using a pretrained depth model would be highly beneficial for our approach.

\medskip

\small


\newpage
\appendix

\section{Geo-localization with Standard Datasets}
We evaluated our method on the standard cross-view datasets CVUSA and CVACT where the orientation is aligned, and the ground map is a panoramic image. In Table~\ref{Standard Cross-View Geo-Localization}, we compare our method with some state-of-the-art methods, including SAFA~\cite{shi2019spatial}, DSM~\cite{shi2020looking}, L2LTR~\cite{yang2021cross}, TransGeo~\cite{zhu2022transgeo}, DeHi~\cite{wang2023dehi}, GeoDTR~\cite{zhang2023cross}, SAIG-D~\cite{zhu2023simple} and Sample4Geo~\cite{deuser2023sample4geo}. It's worth noting that the performance of methods other than ours is taken directly from their original papers, including Sample4Geo. Our method is competitive with the state-of-the-art approaches. Particularly on the larger test dataset, CVACT\_test, our method has even surpassed the state-of-the-art approach, improving from 71.51\% to 72.03\%(+0.53\%) on the recall top-1 metric. 

\begin{table}[h]
	\caption{The performance comparison with state-of-the-art methods on the standard CVUSA and CVACT datasets, where the orientation is aligned, and the ground map is a panoramic image. $\dagger$ denotes which models are using the polar transformation.}
	\label{Standard Cross-View Geo-Localization}
	\centering  \scalebox{0.75}{
		\begin{tabular}{l|cccc|cccc|cccc}
			\toprule
			\multirow{2}*{Approach} & \multicolumn{4}{c}{CVUSA} & \multicolumn{4}{c}{CVACT\_val} & \multicolumn{4}{c}{CVACT\_test} \\
			\cline{2-13}
			& R@1 & R@5 & R@10 & R@1\% & R@1 & R@5 & R@10 & R@1\% & R@1 & R@5 & R@10 & R@1\% \\
			\midrule
			$\dagger$SAFA~\cite{shi2019spatial}       & 89.84 & 96.93 & 98.14 & 99.64 & 81.03 & 92.80 & 94.84 & 98.17 & - & - & - & - \\
			$\dagger$DSM~\cite{shi2020looking}        & 91.96 & 97.50 & 98.54 & 99.67 & 82.49 & 92.44 & 93.99  & 97.32 & - & - & - & - \\
			$\dagger$L2LTR ~\cite{yang2021cross}     & 94.05 & 98.27 & 98.99 & 99.67 & 84.89 & 94.59 & 95.96 & 98.37 & 60.72 & 85.85 & 89.88 & 96.12 \\	
			TransGeo~\cite{zhu2022transgeo}           & 94.08 & 98.36 & 99.04 & 99.77 & 84.95 & 94.14 & 95.78 & 98.37 & - & - & - & - \\
			$\dagger$DeHi~\cite{wang2023dehi}         & 94.34 & 98.63 & 99.22 & 99.82 & 84.96 & 94.48 & 95.98 & 98.58 & - & - & - & - \\
			$\dagger$GeoDTR~\cite{zhang2023cross}     & 95.43  & 98.86  & 99.34 & 99.86 & 86.21 & 95.44 & 96.72 & 98.77 & 64.52 & 88.59 & 91.96 & 98.74\\	
			$\dagger$ SAIG-D~\cite{zhu2023simple}     & 96.34  & 99.10 & 99.50 & 99.86 & 89.06 & 96.11 & 97.08 & 98.89 & 67.49 & 89.39 & 92.30 & 96.80 \\
			Sample4Geo~\cite{deuser2023sample4geo}    & \textbf{98.68} & \textbf{99.68} & \textbf{99.78} & 99.87 & \textbf{90.81} & \textbf{96.74} & \textbf{97.48} & \textbf{98.77} & 71.51 & \textbf{92.42} & \textbf{94.45} & \textbf{98.70} \\
			\textbf{Ours}                                      & 98.65 & 99.67 & 99.75 & \textbf{99.90} & 89.31 & 95.77 & 96.80 & 98.57 & \textbf{72.03} & 91.55 & 93.55 & 98.61\\
			\bottomrule
	\end{tabular}}
\end{table}


\begin{figure}[h]
\centering
\includegraphics[width=1.0\textwidth]{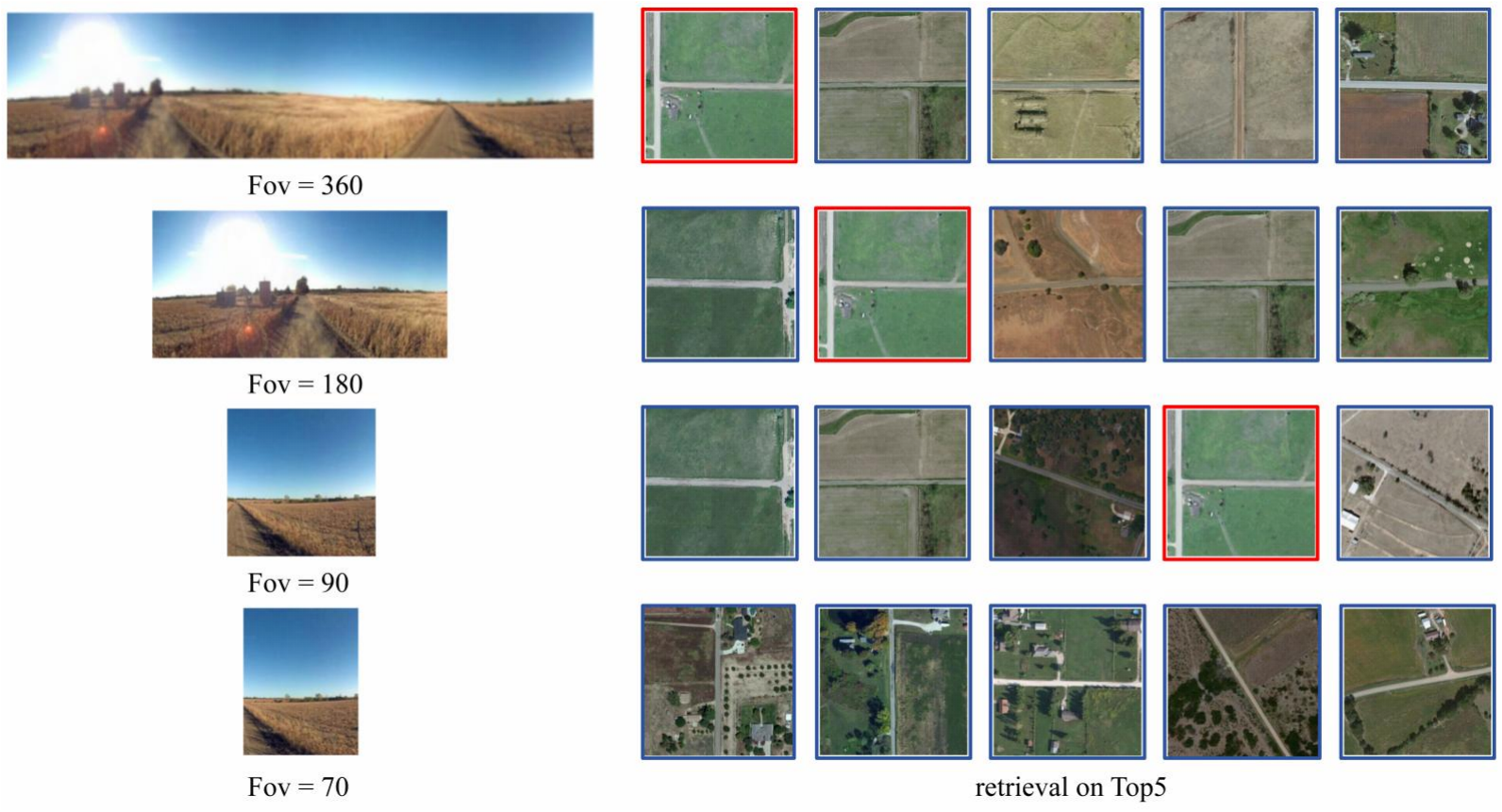}
\caption{Illustration of effect of different FoVs on retrieval. The first column shows the ground pictures with unknown direction and limited FoV. The second to sixth columns show the top-5 pictures in recall, where the red box represents the positive sample and the blue boxes represent the negative samples.}
\label{fig:vary}
\end{figure}	

\section{Qualitative analysis of the impact of different perspectives.}
To more intuitively demonstrate the challenges of geographic localization in real-world scenarios with unknown directions and limited perspectives, we provide examples showcasing how different perspectives affect retrieval results. As shown in the Figure~\ref{fig:vary}, the first column shows the ground pictures with unknown direction and limited FoV. The second to sixth columns show the top-5 pictures in recall, where the red box represents the positive sample and the blue boxes represent the negative samples. We can observe that as the Fov of the ground images decreases, successfully retrieving matching samples becomes more challenging.

Additionally, we add visualization of the aerial image features corresponding to the regions of interest in the ground and aerial images. In each subplot, the first row displays the visualizations of the BEV representations in the regions of interest on the ground and aerial images, while the second row shows the aerial image features. We can observe that the learned BEV representations and the aerial image features focus on highly overlapping regions, effectively reducing the discrepancies between different perspectives.

\begin{figure}[!t]
	\centering
	\subfloat[FoV = 360]{
		\includegraphics[scale=0.3]{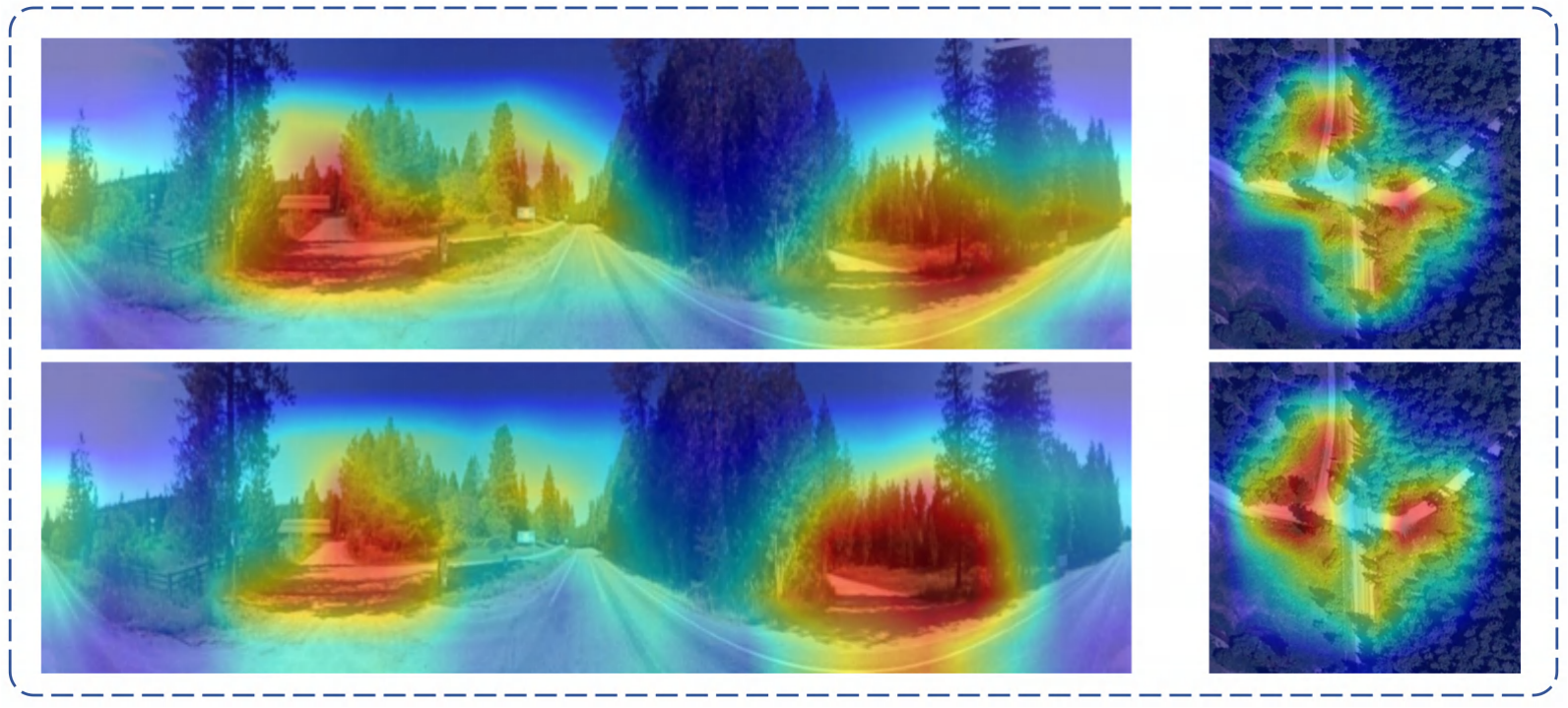}}
	\subfloat[FoV = 90]{
		\includegraphics[scale=0.2]{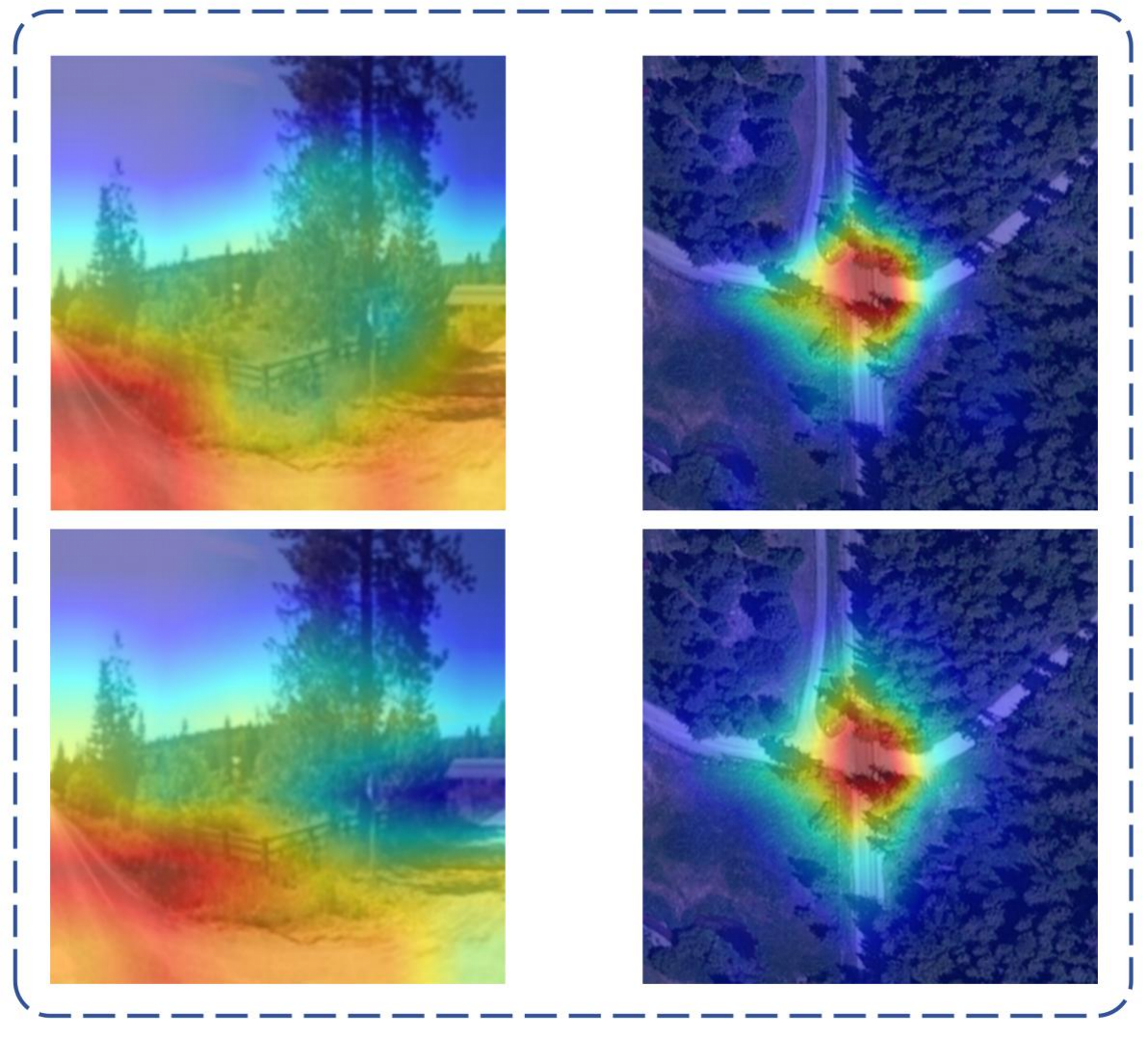}}
	\\
	\subfloat[FoV = 180]{
		\includegraphics[scale=0.39]{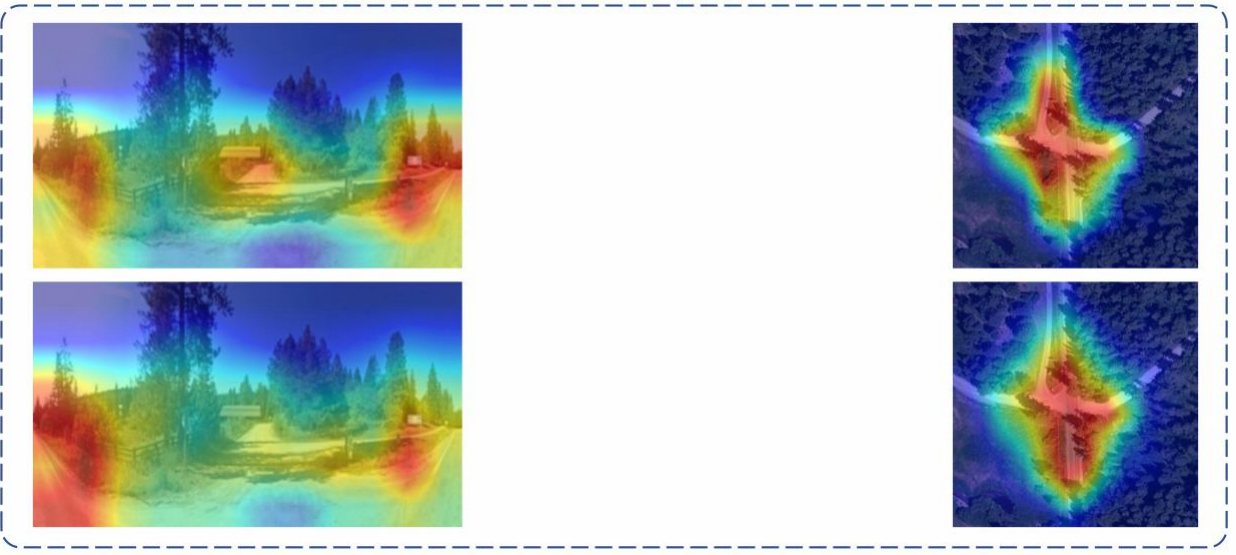}}
	\subfloat[Fov = 70]{
		\includegraphics[scale=0.2]{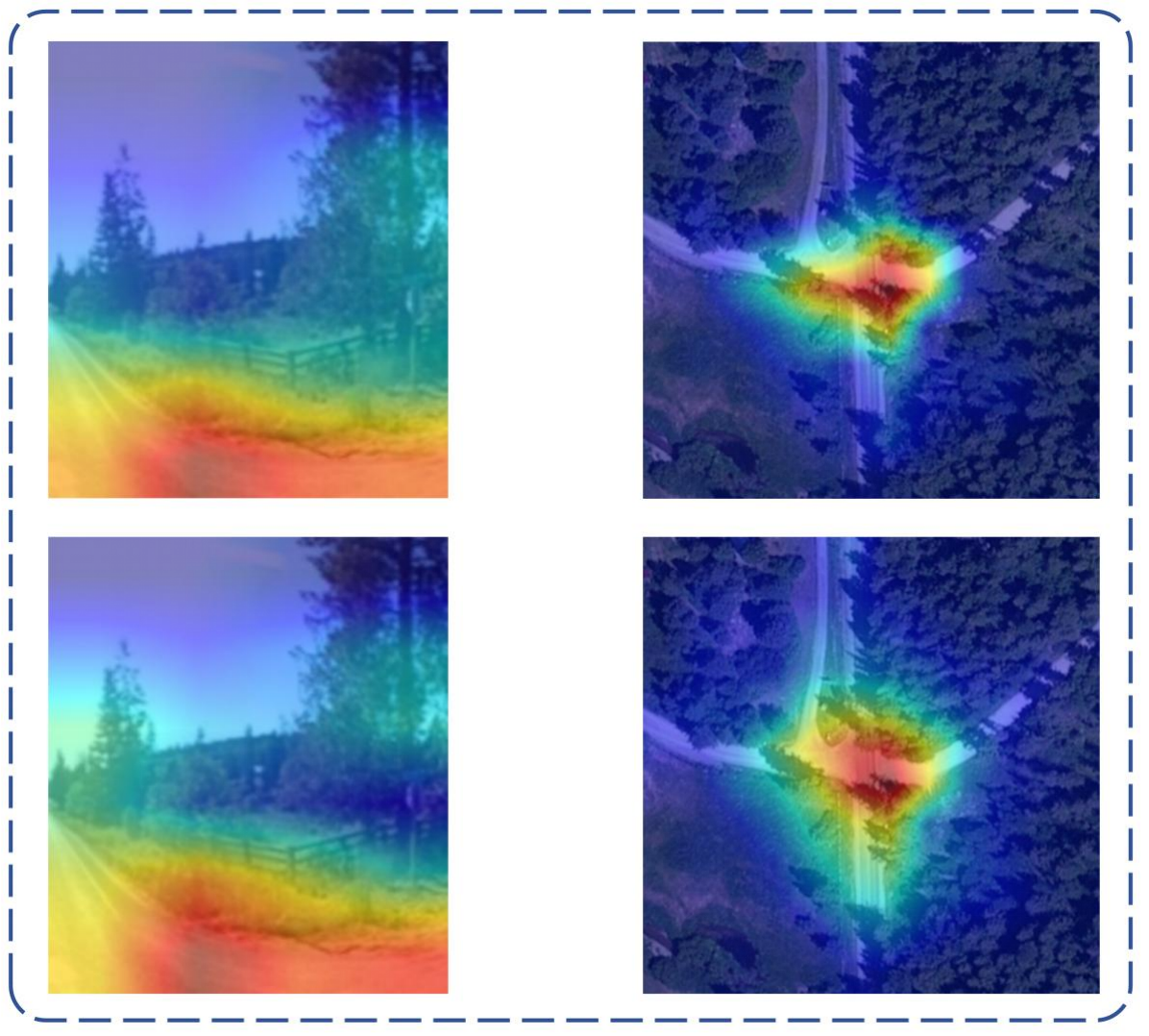}}
	\caption{Illustration of visualization of the BEV representations and aerial-level features corresponding to the regions of interest in ground and aerial images. In each subplot, the first row displays the visualizations of the BEV representations in the regions of interest on the ground and aerial images, while the second row shows the aerial image features.}
	\label{vis}
\end{figure}


\end{document}